\documentclass[10pt, letterpaper, journal, twoside]{IEEEtran}

\IEEEoverridecommandlockouts    
   
\usepackage{graphics} 
\usepackage{cite}
\usepackage[T1]{fontenc}
\usepackage{amsmath, bm} 
\usepackage{amssymb}  
\usepackage{hyperref}
 \hypersetup{
     colorlinks=true,
     linkcolor=blue,
     filecolor=magenta,      
     urlcolor=blue,
 }

\usepackage{float}
\usepackage{cancel}

\usepackage{numprint}
\usepackage{comment}

\usepackage{color, colortbl}
\definecolor{LightCyan}{rgb}{0.88,1,1}
\usepackage{multirow}
\usepackage{siunitx}
\DeclareSIUnit\cell{cell}
\DeclareSIUnit\cells{cells}
\DeclareSIUnit\trees{trees}

\usepackage[nodisplayskipstretch]{setspace}
\usepackage{titlesec} 
\renewcommand{\thesection}{\Roman{section}}
\renewcommand{\thesubsection}{\thesection-\Alph{subsection}}
\renewcommand{\thesubsubsection}{\thesubsection\arabic{subsubsection}}
\titleformat{\subsubsection}[runin]{\itshape}{\arabic{subsubsection})}{0.5em}{}

\titlespacing*{\subsubsection}{\parindent}{0pt}{*1}
\titlespacing*{\section}{0pt}{*1}{*1}
\titlespacing{\subsection}{0pt}{*1}{*1}

\usepackage{enumitem} 

\usepackage{wrapfig}
\usepackage[table]{xcolor}
\usepackage{amssymb}
\usepackage{tikz}
\usetikzlibrary{shapes,arrows}
\usepackage{verbatim}
\usetikzlibrary{positioning}
\usepackage[normalem]{ulem} 
\usepackage{tabstackengine}
\usepackage{algorithm}
\usepackage[end]{algpseudocode} 
\usepackage{etoolbox}
\usepackage[font=small, skip=4pt]{caption}
\usepackage[skip=0pt]{subcaption} 

\usepackage{pgfplots}
\usepackage{grffile}
\pgfplotsset{compat=newest}
\usetikzlibrary{plotmarks}
\usetikzlibrary{arrows.meta}
\usepgfplotslibrary{patchplots}
\usepackage{amsmath}

\usepackage{bbding} 

\usepackage{colortbl}
\usepackage[table]{xcolor}

\usepackage{hhline}
\usepackage{makecell}

\newcommand{\ihab}[1]{{\textcolor{black}{#1}}}

\title{\LARGE \bf Autonomous Navigation of \textit{AGV}s in Unknown Cluttered Environments: \textit{log-MPPI} Control Strategy}

\author{Ihab S. Mohamed$^{1}$, Kai Yin$^{2}$, and Lantao Liu$^{1}$
\thanks{Manuscript received: February 24, 2022; Accepted: July 12, 2022. 
This letter was recommended for publication by Editor Ashis Banerjee upon evaluation of the Associate Editor and Reviewers' comments.
}
\thanks{$^{1}$Ihab S. Mohamed (\ihab{corresponding author}) and Lantao Liu are with the Luddy School of Informatics, Computing, and Engineering, Indiana University, Bloomington, IN 47408 USA, {\tt\small \{mohamedi, lantao\}@iu.edu}}
\thanks{$^{2}$Kai Yin is with Expedia Group, USA, 
    {\tt\small kyin@expediagroup.com}
}
\thanks{ Digital Object Identifier (DOI): see top of this page.}
}%
\definecolor{applegreen}{rgb}{0.8, 1, 0.0}
\definecolor{LightCyan}{rgb}{0.88,1,1}
\definecolor{atomictangerine}{rgb}{1.0, 0.6, 0.4}
\definecolor{amber}{rgb}{1.0, 0.75, 0.0}
\definecolor{aqua}{rgb}{0.0, 1.0, 1.0}
\definecolor{almond}{rgb}{0.94, 0.87, 0.8}
\definecolor{aquamarine}{rgb}{0.5, 1.0, 0.83}
\definecolor{babyblue}{rgb}{0.54, 0.81, 0.94}
\definecolor{babyblueeyes}{rgb}{0.63, 0.79, 0.95}
\definecolor{asparagus}{rgb}{0.53, 0.66, 0.42}
\definecolor{auburn}{rgb}{0.43, 0.21, 0.1}
\definecolor{brilliantlavender}{rgb}{0.96, 0.73, 1.0}
\definecolor{bittersweet}{rgb}{1.0, 0.44, 0.37}
\definecolor{blue-violet}{rgb}{0.54, 0.17, 0.89}
\definecolor{capri}{rgb}{0.0, 0.75, 1.0}
\definecolor{celadon}{rgb}{0.67, 0.88, 0.69}
\definecolor{darkcyan}{rgb}{0.0, 0.55, 0.55}
\definecolor{deepskyblue}{rgb}{0.0, 0.75, 1.0}
\definecolor{dogwoodrose}{rgb}{0.84, 0.09, 0.41}

\begin{document}

\maketitle

\global\csname @topnum\endcsname 0
\global\csname @botnum\endcsname 0




\begin{abstract}
Sampling-based model predictive control (\textit{MPC}) optimization methods, such as Model Predictive Path Integral (\textit{MPPI}), have recently shown promising results in various robotic tasks. However, it might produce an {infeasible} trajectory when the distributions of all sampled trajectories are concentrated within high-cost even infeasible regions. 
In this study, we propose a new method called \textit{log-MPPI} equipped with a more effective trajectory sampling distribution policy which significantly improves the trajectory {feasibility} in terms of satisfying system constraints. 
The key point is to draw the trajectory samples from the normal log-normal (\textit{NLN}) mixture distribution, rather than from Gaussian distribution. 
Furthermore, this work presents a method for collision-free navigation in unknown cluttered environments by incorporating the \textit{2D} occupancy grid map into the optimization problem of the sampling-based \textit{MPC} algorithm.
We first validate the efficiency and robustness of our proposed control strategy through extensive simulations of \textit{2D} autonomous navigation in different types of cluttered environments as well as the cartpole swing-up task. 
We further demonstrate, through real-world experiments, the applicability of \textit{log-MPPI} for performing a \textit{2D} grid-based collision-free navigation in an unknown cluttered environment, showing its superiority to be utilized with the local costmap without adding additional complexity to the optimization problem. 
\ihab{A video demonstrating the real-world and simulation results is available at \url{https://youtu.be/_uGWQEFJSN0}}.


\end{abstract}


\vspace*{-5pt}
\section{Introduction and Related Work}\label{Introduction}
Designing a safe, reliable, and robust control methodology for autonomous navigation of autonomous ground vehicles (\textit{AGVs}) in unknown cluttered environments (such as dense forests, crowded offices, corridors, warehouses, etc.) has been known as a great challenge. 
Such a navigation task requires the \textit{AGV} to navigate safely with full autonomy while avoiding getting trapped in local minima and collision with static and dynamic obstacles while moving towards the goal, as well as respecting various system constraints. 
To this end, the robot should be capable of perceiving 
its surrounding environment and then reacting adequately.
This subsequently results in a complex optimization control problem that is difficult to be solved in \textit{real-time} \cite{mohamed2020model}. 
\begin{figure}%
    \centering
    \subfloat[Indoor validation-used environment]{\vspace*{1pt}{\includegraphics[scale=1]{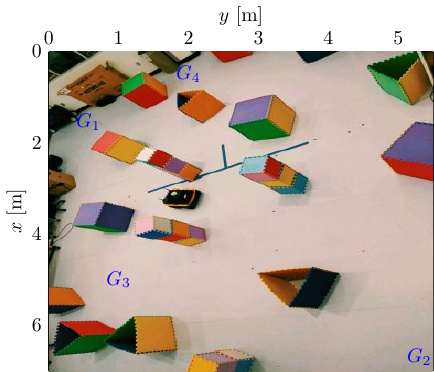}}\label{fig:real-world-enviro}}%
    \;
    \subfloat[\textit{2D} costmap visualization]{\vspace*{1pt}{\includegraphics[height=1.3in, width=.45\columnwidth]{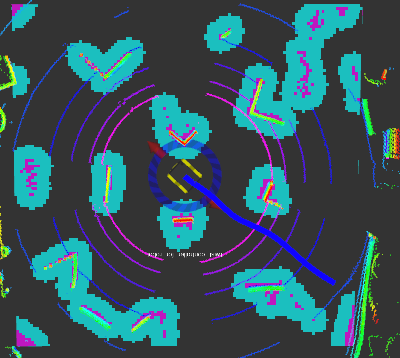}}\label{fig:indoor-costmap}}%
    \caption{Snapshot of (a) our Jackal robot equipped with a Velodyne VLP-16 LiDAR and located in an unknown cluttered environment {($G_1,\dots, G_4$: desired poses)}, and (b) the corresponding \textit{2D} local costmap that represents the surrounding obstacles, where \textcolor{purple}{purple} and \textcolor{cyan}{cyan} cells represent obstacles and their inflation, respectively; while \textcolor{blue}{blue} line is the planned trajectory by our \textit{log-MPPI} control strategy.
    }%
    \label{fig:indoor-cluttered-enviroment}%
    \vspace*{-12pt} 
\end{figure}

One of the well-established and promising control strategies for collision-free navigation is Model Predictive Control (\textit{MPC}) strategy, owing to its flexibility and ability to compute 
good control policy in the presence of the hard and soft constraints of the system to be controlled. It leverages a receding horizon strategy to plan a sequence of optimal control inputs over a prediction time-horizon; then, the first control input in the sequence is applied to the system, while the remaining control sequence is used for warm-starting the optimization at the next time-step \cite{mayne2000constrained}. The existing \textit{MPC}-based schemes can be mainly categorized into gradient-based and sampling-based trajectory optimization methods. 
The gradient-based \textit{MPC} methods have been successfully applied to real robotic systems, obtaining smooth collision-free trajectories in the presence of obstacles and other constraints \cite{gaertner2021collision, lindqvist2020nonlinear, brito2019model, abbas2017obstacle}.
However, the gradient-based frameworks are typically based on strong assumptions: the cost function, and sometimes system constraints, 
need to be differentiable in order to leverage the gradient for computing the optimal solution. 
Unfortunately, in practice, the optimization problem often involves non-convex and non-differentiable objective function with discontinuous collision avoidance constraints, and it can be computationally difficult to be solved in \textit{real-time}.
To alleviate these issues, for example, the non-differentiable system constraints can be reformulated into smooth and differentiable ones as presented in \cite{zhang2020optimization}. 
A promising alternative is the sampling-based optimization methods such as Model Predictive Path Integral (\textit{MPPI}) control strategy \cite{williams2017model} that (i) makes no assumptions or approximations on the objective functions and system dynamics, (ii) can be effectively applied on highly dynamic systems, and (iii) benefits from the parallel nature of sampling and high computational capacities of Graphics Processing Units (\textit{GPUs}) utilized for speeding up the optimization. 

Recently, \textit{MPPI} or sampling-based \textit{MPC} has been successfully applied to a wide variety of robotic applications, starting from aggressive driving \cite{williams2016aggressive, williams2018information} and autonomous flight \cite{mohamed2020model, lu2022real} and ending with visual servoing \cite{mohamed2021sampling} and reactive manipulation 
\cite{bhardwaj2022storm}, showing outstanding performance in the presence of non-convex and discontinuous objectives, without adding any additional complexity to the optimization problem.
Despite the attractive characteristics of \textit{MPPI} \cite{mohamed2021mppi}, much like any sampling-based optimization method, it might generate an \textit{infeasible} control sequence (i.e., \textit{infeasible} trajectory) when the distributions of all sampled trajectories drawn from the system dynamics are 
unfortunately surrounding some \textit{infeasible} 
region. This may inevitably lead to either violating the system constraints or increasing the risk of being trapped in local minima. 
For tasks such as collision-free navigation in cluttered environments, 
it has been observed 
that the collisions may not be avoided, as the intensive simulations demonstrated in Section~\ref{Simulation Details and Results}. 
To mitigate this problem, authors in \cite{williams2018robust} employed an iterative Linear Quadratic Gaussian (\textit{iLQG}) control, as an ancillary controller, on top of \textit{MPPI} for tracking the planned trajectory. 
Similarly, in \cite{pravitra2020L}, \textit{MPPI} is augmented with a nonlinear $\mathcal{L}_1$ adaptive controller to compensate for the model uncertainty.
Lately, the covariance steering (\textit{CS}) principle is incorporated within the \textit{MPPI} algorithm, aiming to introduce adjustable trajectory sampling distributions \cite{yin2021improving}. 
However, the proposed solutions in \cite{pravitra2020L} and \cite{yin2021improving} have not been experimentally validated on real robotic systems.

\vspace*{-0.3pt}
With the aim of mitigating the previous shortcomings and improving the performance of \textit{MPPI}, we propose a new strategy, called \textit{log-MPPI} control strategy, that provides more flexible and efficient distributions of the sampled trajectories\ihab{, without the need for (i) integrating an ancillary controller on top of \textit{MPPI} \cite{pravitra2020L, williams2018robust}, or (ii) adding a feedback term along with the injected artificial noise that requires the system dynamics to be linearized and converts the original optimization problem into a convex one \cite{yin2021improving}}.
The key idea of \textit{log-MPPI} is that the trajectories (or, the control input updates) are sampled from the product of normal and log-normal distribution (namely, \textit{NLN} mixture), instead of sampling from only Gaussian distribution. 
{With such a sampling strategy, a small injected noise variance can be utilized so that violating system constraints can be avoided; yet, we can still get desirable sampled rollout trajectories that are well spread-out for covering large state-space.}
In summary, the contributions of this work can be summarized as follows:
\begin{enumerate}
    \item We provide a new sampling strategy based on {normal} log-normal {mixture} distribution. This new sampling method  provides more efficient trajectories than the vanilla \textit{MPPI} variants, ensuring a much better exploration of the state-space of the given system and reducing the risk of getting stuck in local minima, leading  the robot to ultimately find {feasible} trajectories that avoid  collision, as revealed in Section~\ref{Task1:Cluttered Environments}. 
    \item Then, we show through the cartpole swing-up task given in Section~\ref{Cart-Pole Swing-up Task} the robustness of \textit{log-MPPI} to run with a significantly fewer number of trajectories, opening up a new avenue for the sampling-based \textit{MPC} algorithm to be run on a standard \textit{CPU} instead of a \textit{GPU}.
    We thereafter incorporate the \textit{2D} grid map, as a local costmap, into the sampling-based \textit{MPC} algorithm for performing collision-free navigation in either static or dynamic unknown cluttered environments, as depicted in Section~\ref{Task2: Unknown Environments}.
    \item Finally, in Section~\ref{Real-World Demonstration}, we experimentally demonstrate our control strategy for a 2D grid-based navigation in an unknown indoor cluttered environment shown in Fig.~\ref{fig:indoor-cluttered-enviroment}; 
    to the best of the authors' knowledge, this is the first attempt to experimentally achieve grid-based collision-free navigation based on sampling-based \textit{MPC}.
\end{enumerate}
\section{Preliminaries}\label{preliminaries}
In this section, we define the optimal control problem to be solved and provide a brief review of \textit{MPPI}.
\vspace{-3pt}
\subsection{Constrained Control Problem}
Consider a discrete-time system with state $\mathbf{x}_k \in \mathbb{R}^{n}$, control input $\mathbf{u}_k \in \mathbb{R}^{m}$, and underlying dynamics $\mathbf{x}_{k+1}=f\left(\mathbf{x}_{k},\mathbf{v}_{k}\right)$.
Let $\mathbf{v}_{k} = \mathbf{u}_{k}+\delta \mathbf{u}_{k}\sim \mathcal{N}(\mathbf{u}_k, \Sigma_{\mathbf{u}})$ where $\delta \mathbf{u}_{k} \sim \mathcal{N}(\mathbf{0}, \Sigma_{\mathbf{u}})$ represents the injected disturbance with a zero-mean and co-variance $\Sigma_{\mathbf{u}}$.
Let $\mathbf{U} = \left(\mathbf{u}_{0}, \mathbf{u}_{1}, \dots,\mathbf{u}_{N-1}\right) \in \mathbb{R}^{m \times N}$ denote the control sequence over a finite time-horizon $N$, while
$\mathbf{X} = \left(\mathbf{x}_{0}, \mathbf{x}_{1}, \dots, \mathbf{x}_{N-1}\right) \in \mathbb{R}^{n \times N}$ denotes the resulting state trajectory.
Let $\mathcal{O}^{\text {rob}}\left(\mathbf{x}_{k}\right)$ and $\mathcal{O}^{\text {obs}}$ be the area occupied by the robot and the obstacles, respectively.  
Our objective is to find a control sequence $\mathbf{U}$ that generates a collision-free trajectory which allows the robot to navigate from the initial state $\mathbf{x}_s$ to its desired state $\mathbf{x}_f$ while minimizing a cost function $J$.
This optimization problem can be formulated by \textit{MPPI} as:
\vspace*{-3pt}
\begin{subequations}
\begin{align}
\min _{\mathbf{U}} \quad  J &=  \mathbb{E}\left[\phi\left(\mathbf{x}_{N}\right)+\sum_{k=0}^{N-1}\left(q\left(\mathbf{x}_{k}\right)+\frac{1}{2} \mathbf{u}_{k}^{T} R \mathbf{u}_{k}\right)\right]\!, \label{eq:2a}\\
\text {s.t.} \quad & \mathbf{x}_{k+1}=f\left(\mathbf{x}_{k}, \mathbf{v}_{k}\right), \label{eq:2b}\\
& \mathcal{O}^{\text {rob}}\left(\mathbf{x}_{k}\right) \cap \mathcal{O}^{\text {obs}}=\emptyset, \;\mathbf{h}(\mathbf{x}_k, \mathbf{u}_k) \leq 0, \label{eq:2c}\\
& \mathbf{x}_0 = \mathbf{x}_s, \;\mathbf{u}_{k} \in \mathbf{U},\; \mathbf{x}_{k} \in \mathbf{X}, 
\end{align}
\end{subequations}
where $\phi\left(\mathbf{x}_N\right)$, $q\left(\mathbf{x}_{k}\right)$, and $R\in \mathbb{R}^{m \times m}$ denote the terminal cost, state-dependent running cost, and positive definite control weighting matrix, respectively. \textit{MPPI} solves the problem by minimizing the objective (\ref{eq:2a}) subject to system dynamics (\ref{eq:2b}) and system constraints such as collision avoidance and control constraints (\ref{eq:2c}). 
\subsection{Review of MPPI}
Unlike the gradient-based \textit{MPC} methods, \textit{MPPI} does not compute gradients to find the optimal solution; i.e., it is a derivative-free trajectory optimization strategy. Moreover, it makes no assumptions on the objective functions and system dynamics; i.e., highly nonlinear and non-convex functions can be easily employed.
At each time-step, \textit{MPPI} draws $M$ trajectories, in parallel, from the system dynamics using \textit{GPU} ensuring a \textit{real-time} performance.
These parallel trajectories are then evaluated according to its expected cost.
The \textit{cost-to-go} of each rollout $\tau$ over a time-horizon $N$ is given by
\vspace*{-8pt}
\begin{equation}
\vspace*{-4pt}
 \tilde{S}\left(\tau\right) =\phi\left(\mathbf{x}_N\right)+\sum_{k=0}^{N-1} \tilde{q}\left(\mathbf{x}_{k}, \mathbf{u}_{k}, \delta \mathbf{u}_{k}\right),
\end{equation}
where $\tilde{q}\left(\mathbf{x}_k, \mathbf{u}_k, \delta \mathbf{u}_k\right)$ refers to the instantaneous running cost which consists of the sum of state-dependent running cost $q\left(\mathbf{x}_{k}\right)$ and quadratic control cost; it is defined as \cite{williams2017model} 
\begin{equation}\label{eq:cost-to-go}
\tilde{q} 
\!= \!
\underbrace{
\vphantom{
\frac{\left(1\!-\!\frac{1}{\nu}\right)}{2} \delta \mathbf{u}_{k}^{T} R \delta \mathbf{u}_{k}+\mathbf{u}_{k}^{T} R \delta \mathbf{u}_{k}+\frac{1}{2} \mathbf{u}_{k}^{T} R \mathbf{u}_{k}}
q\left(\mathbf{x}_{k}\!\right)}_{\text{\color{black}{\textit{State-dep.}}}} 
\!+ \!
\underbrace{\frac{\left(1\!-\!\nu^{-1}\right)}{2} \delta \mathbf{u}_{k}^{T} R \delta \mathbf{u}_{k}\!+ \!\mathbf{u}_{k}^{T} R \delta \mathbf{u}_{k}\!+ \!\frac{1}{2} \mathbf{u}_{k}^{T} R \mathbf{u}_{k}}_{\text{\color{black}{\textit{Quadratic Control Cost}}}},
\end{equation}
where $\nu \in \mathbb{R}^{+}$ determines how aggressively the state-space is explored.
Afterwards, the control sequence is updated based on a weighted average cost over all sampled trajectories.
As described in \cite{williams2017model}, the optimal control sequence $\left\{\mathbf{u}_{k}\right\}_{k=0}^{N-1}$ can be approximated
as
\begin{equation}\label{eq:mppi_optimal-control}
 \mathbf{u}_{k+1} = \mathbf{u}_{k} +\frac{\sum_{m=1}^{M} \exp \left(-(1 / \lambda) \tilde{S}\left(\tau_{k, m}\right)\right) \delta \mathbf{u}_{k, m}}{\sum_{m=1}^{M} \exp \left(-(1 / \lambda) \tilde{S}\left(\tau_{k, m}\right)\right)},
 \end{equation}
where $\tilde{S}\left(\tau_{k, m}\right) \in \mathbb{R}^{+}$ is the \textit{cost-to-go} of the $m^{th}$ trajectory at $k^{th}$ step and $\lambda \in \mathbb{R}^{+}$ is so-called the inverse temperature which determines how selective the weighted average of the trajectories is \cite{williams2018information}.  
The control sequence is then smoothed using a Savitzky-Galoy (SG) filter. 
Finally, the first control $\mathbf{u}_{0}$ in the sequence is applied to the system, while the remaining control sequence of length $N-1$ is used for warm-starting the optimization at the next time-step.
\section{\textit{log-MPPI} Control Strategy}\label{log-MPPI Control Strategy}
Our goal is to design a new sampling and control approach, the \textit{log-MPPI}, to further improve the classic \textit{MPPI} performance. 
{
Here, we briefly describe the difference between \textit{MPPI} and \textit{log-MPPI}: 
as previously discussed in Section~\ref{preliminaries}, \textit{MPPI} does not update the injected control noise variance $\Sigma_{\mathbf{u}}$, and the state-space exploration is carried out by adjusting $\nu$ (see Eq.~\eqref{eq:cost-to-go}). However, if $\nu$ is too large, \textit{MPPI} produces control inputs with significant chatter \cite{williams2017model}. Similarly, as stated in \cite{williams2018robust}, a higher value of $\Sigma_{\mathbf{u}}$ might result in violating the system constraints and eventually state diverging. One solution could be updating, at each iteration, the variance \cite{bhardwaj2022storm}. Instead, we inject the log-normal along with normal distribution ensuring much better state exploration with a low variance value which well respects the system constraints and providing better performance with a fewer number of samples.}
\subsection{Log-normal Distribution}
In probability theory, a positive random variable $X$ is log-normally distributed, i.e., $X \sim \mathcal{LN}(\mu, \sigma^2)$, if the natural logarithm of  $X$ has a normal distribution with mean $\mu$ and variance $\sigma ^{2}$, i.e., $\ln X \sim \mathcal{N}(\mu, \sigma^2)$. Thus, the probability density function (\textit{pdf}) of the random variable $X$ is given by
\begin{equation}
f_{X}(x)=\frac 1 {x\sigma\sqrt{2\pi\,}} \exp\left( -\frac{(\ln x-\mu)^2}{2\sigma^2} \right), x \in \mathbb{R}^{+}, 
\end{equation}
where the mean and variance of a $\mathcal{LN}(\mu,\sigma^2)$ distribution are given by
\begin{equation}\label{eq:LN_m-var}
\mathbb{E} (X) = e^{(\mu + 0.5\sigma^2)} \; \text{and} \; \operatorname{Var} (X)=e^{(2\mu +\sigma^2)}(e^{\sigma^2} - 1).  
\end{equation}
In spite of the fact that both normal and log-normal distributions are unbounded distributions, the log-normal distribution is asymmetric and positively skewed to the right, where the range of values lies in an interval of $[0,+\infty[$. Therefore, the \textit{pdf}, $f_{X}(x)$, starts at zero and increases to its mode, then decreases thereafter. The degree of skewness increases as $\sigma$ increases, for a given $\mu$. Similarly, for the same $\sigma$, the \textit{pdf}'s skewness increases as $\mu$ increases.  
In addition, the most attractive feature of such distribution compared to the alternative default distributions (such as normal, gamma, and Weibull distributions) is its capability of capturing a large range with a long \textit{right-tail}, making it convenient to model large values and hence large uncertainties \cite{johnson1995continuous}.
\begin{figure}%
    \subfloat[$X\!\sim\! \mathcal{N}(0,0.002)$]{\vspace*{0pt}{\hspace*{-5pt}\includegraphics[scale=1]{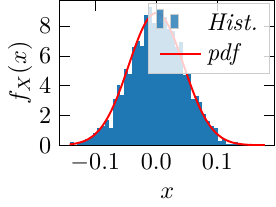}}\label{fig:normal}}%
    \subfloat[$\!\!Y \!\!\!\sim\! \mathcal{LN}(1.023, 0.048)$]{\vspace*{0pt}{\hspace*{0pt}\includegraphics[scale=1]{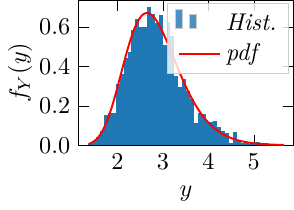}}\label{fig:lognormal}}%
    \subfloat[$Z \!\sim \!\mathcal{NLN}(0, 0.017)$]{\vspace*{0pt}{\includegraphics[scale=1]{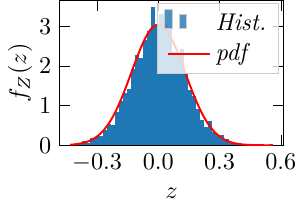}}\label{fig:normal-lognormal}}%
    \caption{Histogram and \textit{pdf} of $2500$ random samples generated from: (a) normal, (b) log-normal, and (c) \textit{NLN} mixture, with $\mu_n = \mu_{nln} = 0, \mu_{ln} = 1.023$, $\sigma^2_n = 0.002, \sigma^2_{ln} = 0.048,$ and $\sigma^2_{nln} = 0.017$.}%
    \label{fig:histogram}%
\vspace*{-12pt} 
\end{figure}

\subsection{Normal Log-normal (\textit{NLN}) Mixture}
Broadly speaking, if $X$ and $Y$ are two independent random variables, described by probability density functions $f_X(x)$ and $f_Y(y)$, then the probability density function of the product $Z= XY$ is given by
\begin{equation}\label{productZ} 
    f_Z(z) = \int^\infty_{-\infty} f_X(x)  f_Y(z/x)  \frac{1}{|x|}\, dx.
\end{equation}
More specifically, suppose that $X\sim \mathcal{N}(\mu_n, \sigma_n^2)$ and $Y\sim \mathcal{LN}(\mu_{ln}, \sigma_{ln}^2)$. Then, the random variable $Z= XY\sim \mathcal{NLN}(\mu_{nln}, \sigma_{nln}^2)$ can be labelled as normal log-normal (\textit{NLN}) mixture, where its mean $\mu_{nln}$ and variance $\sigma_{nln}^{2}$ are given by \cite{rohatgi2015introduction}
\vspace{-3pt}
\begin{equation}\label{eq:mean-var-NLN}
    \begin{aligned}
        \mu_{nln} &= \mathbb{E} (X Y) = \mathbb{E} (X) \mathbb{E} (Y) = \mu_{n} e^{(\mu_{ln}+ 0.5\sigma_{ln}^{2})}, \\
          \sigma_{nln}^{2} &= \operatorname{Var} (X Y) =\mathbb{E} (X^{2}) \mathbb{E} (Y^{2})-(\mathbb{E} X Y)^{2}, \\
            &=\left(\mu_{n}^{2}+\sigma_{n}^{2}\right) e^{(2 \mu_{ln}+2 \sigma_{ln}^{2})}-\mu_{n}^{2} e^{(2 \mu_{ln}+\sigma_{ln}^{2})}.
         \end{aligned}
         \vspace{-3pt}
    \end{equation}
Let us consider the case where $\mu_n=0$, i.e., $X\sim \mathcal{N}(0, \sigma_n^2)$, which is commonly used in sampling-based \textit{MPC} strategies.  
Thus, the \textit{pdf} of $Z$, given in (\ref{productZ}), is defined as 
\begin{align}
f_Z\!= \! \frac{1}{2\pi \sigma_n \sigma_{ln}} \! \!   \int_0^\infty \! \!  \! \! \!\exp \!\left(\!-\frac{z^2}{2\sigma_n^2 x^2} - \frac{(\ln x-\mu_{ln})^2}{2\sigma_{ln}^2} \! \right) \!\frac{1}{x^2} dx,
\end{align}
which can be solved analytically \cite{clark1973subordinated}.
It is noteworthy that $f_Z(z) = f_Z(-z)$, indicating a \textit{symmetric} distribution around $0$ as shown in Fig.~\ref{fig:normal-lognormal}. 
{In addition, $Z$ can be written as a smooth function of two independent normal distributions, 
i.e., $Z= XY=X e^W$, where $W\sim \mathcal{N}(\mu_{ln},\sigma^2_{ln})$. 
When $\sigma_n$ becomes smaller, $X$ places the mass around $0$. This makes the tail of $Z$ lighter than the lognormal distribution. }
Fig.~\ref{fig:histogram} illustrates different distributions with differing parameters.  


\subsection{\textit{log-MPPI} Control Strategy}
{
We develop our method on top of the \textit{MPPI} in \cite{williams2017model} for integrating the \textit{NLN} mixture sampling. Although the original derivation of \textit{MPPI} is based on the controlled dynamics driven by Brownian motion noise, it can be approximately applied to the \textit{NLN} mixture, particularly for small $\sigma_n$. 
We provide a discussion on the effect of \textit{NLN} mixture noise on the dynamics 
in Appendix~\ref{Appendix-dynamics-lognormal-normal}.
}
The major difference is that the trajectories, drawn from the discrete-time dynamics system $\mathbf{x}_{k+1}$, are sampled from the \textit{NLN} policy, rather than from the Gaussian policy. 
Accordingly, the control input updates $\delta \mathbf{u}_{k} $ is defined as $\delta \mathbf{u}_{k}  = \delta \mathbf{u}^n_{k} \cdot \delta \mathbf{u}^{ln}_{k} \sim \mathcal{NLN}(\mathbf{0}, \Sigma_{\mathbf{u}})$, where 
$\delta \mathbf{u}^n_{k} \sim \mathcal{N}(\mathbf{0}, \Sigma_{n})$, $\delta \mathbf{u}^{ln}_{k} \sim \mathcal{LN}(\mu_{ln}, \Sigma_{ln})$, $\Sigma_\mathbf{u} = \sigma^2_{nln}\mathbf{I}_m$, $\Sigma_{n} = \sigma^2_{n}\mathbf{I}_m$, $\Sigma_{ln} = \sigma^2_{ln}\mathbf{I}_m$, and $ \mathbf{I}_m$ denotes an $m \times m$ identity matrix.
To ensure that $\delta \mathbf{u}^{ln}_{k}$ is \textit{stochastically} independent from $\delta \mathbf{u}^{n}_{k}$, Eq. (\ref{eq:LN_m-var}) is employed in order to compute $\mu_{ln}$ and $\Sigma_{ln}$, considering $\mu = 0$ and 
$\sigma^2 = \sigma_{n}$ (namely, standard deviation of $\delta \mathbf{u}^{n}_{k}$).  
Similarly, Eq. (\ref{eq:mean-var-NLN}) is utilized for computing $\mu_{nln}$ and $\sigma^2_{nln}$, taking into account that $\mu_{nln} = 0$ as $\mu_{n} = 0$.
\begin{figure}[!ht]%
\vspace*{-10pt}
    \centering
    \subfloat[$\delta \mathbf{u}_{k} \sim \mathcal{N}(\mathbf{0},0.002\mathbf{I}_2)$]{\vspace*{-1pt}{\includegraphics[scale=0.8]{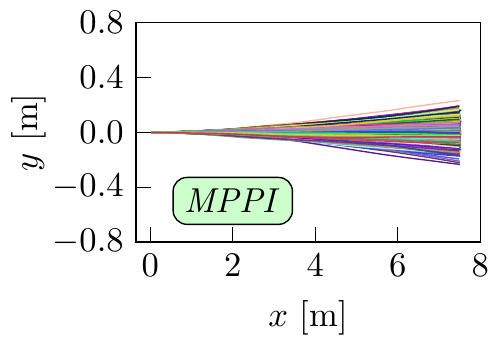}}\label{fig:samples-normal}}%
    \quad
    \subfloat[$\!\delta \mathbf{u}_{k} \!\sim \!\mathcal{NLN}(\mathbf{0}, 0.017\mathbf{I}_2)$]{\vspace*{-1pt}{\includegraphics[scale=0.8]{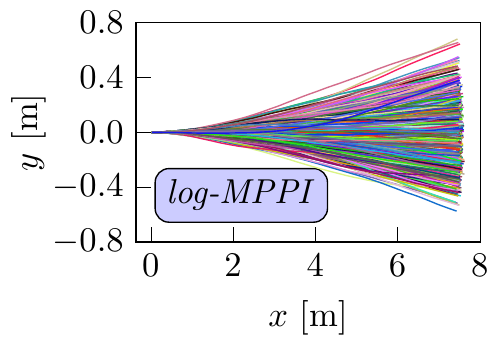}}\label{fig:samples-normal-lognormal}}%
    \caption{Distribution of $2500$ sampled trajectories generated by (a) \textit{MPPI} with $\delta \mathbf{u}_{k} \sim \mathcal{N}(\mathbf{0}, 0.002\mathbf{I}_2)$ and (b) \textit{log-MPPI} with $\delta \mathbf{u}_{k} \sim \mathcal{NLN}(\mathbf{0}, 0.017\mathbf{I}_2)$, where the robot is assumed to be located at $\mathbf{x} = [{x}, {y}, \theta]^T= [0,0,0]^T$ in ([\si{\metre}], [\si{\metre}], [\si{\deg}]) with a commanded control input $\mathbf{u} = [v,\omega]^T = [1.5, 0]^T$ in ([\si{\metre/\second}], [\si{
\radian/\second}]).}%
    \label{fig:generated-samples}%
    \vspace*{-3pt}
\end{figure}

To demonstrate the advantages of sampling from the \textit{NLN} distribution on the performance of the \textit{MPPI} algorithm, here we provide a concrete example. 
The basic idea is to use a small variance (so that we can avoid violating system constraints); yet, we can still get desirable sampled trajectories that are well spread-out for covering large state-space. 
Specifically, 
(i) we first draw random samples, namely, $X \equiv \delta \mathbf{u}^{n}_{k}$, from a normal distribution with $\mu_{n} = 0$ and $\sigma^2_{n}=0.002$ (see Fig.~\ref{fig:normal}); (ii) then, another set of random samples, namely, $Y \equiv \delta \mathbf{u}^{ln}_{k}$, are generated from the corresponding log-normal distribution with $\mu_{ln} =1.023$ and $\sigma^2_{ln}=0.048$, as illustrated in Fig.~\ref{fig:lognormal}; (iii) finally, the random variable $Z \equiv \delta \mathbf{u}_{k}$ that represents the product of those two independent variables is generated from the \textit{NLN} distribution with $\mu_{nln} = 0$ and $\sigma^2_{nln}=0.017$.
Now, let us draw $M$ sampled trajectories from $\delta \mathbf{u}_{k} \sim \mathcal{N}(\mathbf{0}, 0.002\mathbf{I}_2)$, as depicted in Fig.~\ref{fig:samples-normal}, considering the discrete-time kinematics model of the robot given in (\ref{eq:kinematics-model}) and {control schemes} parameters listed in Section~\ref{Simulation Setup:Cluttered Environments}, where $M=2500$. In a similar way, Fig.~\ref{fig:samples-normal-lognormal} shows the sampled trajectories from $\delta \mathbf{u}_{k} \sim \mathcal{NLN}(\mathbf{0}, 0.017\mathbf{I}_2)$.
It is interesting to observe in Fig.~\ref{fig:generated-samples} that the distributions of the sampled trajectories generated by the \textit{log-MPPI} algorithm are more flexible and efficient than the ones generated by the classical \textit{MPPI}, resulting in (i) better exploration of the state-space, and (ii) reducing the probability of getting trapped in local minima.

One might argue that injecting the same control variance to the normal distribution can lead to  similar results. \ihab{Here, we provide the advantages of the \textit{NLN} distribution through (i) an analysis from the dynamics perspective in Appendix~\ref{Appendix-dynamics-lognormal-normal} where we show that even with the same variance, the proposed scheme can be more efficient due to the random drift term in dynamics, and (ii) the extensive simulation results carried out in the next section which show a much better exploration with more than 30\% reduction in the injected noise variance $\Sigma_\mathbf{u}$\footnote{\ihab{Unlike in \cite{yin2021improving}, the proposed method explores the environment and samples trajectories more efficiently than \textit{MPPI} for the same injected noise $\Sigma_\mathbf{u}$.}}.}

\section{Simulation-Based Evaluation}\label{Simulation Details and Results}
{We evaluated and compared the two control strategies on a simulated cartpole system and a goal-oriented \textit{AGV} autonomous navigation task in \textit{2D} cluttered environments.} 
\subsection{Cartpole Swing-up Task}\label{Cart-Pole Swing-up Task}
To demonstrate the impact of drawing trajectories from the \textit{NLN} distribution policy, instead of Gaussian policy, on the behavior of the sampling-based \textit{MPC} algorithm \ihab{and assess the \textit{practical} stability of our proposed control strategy, especially with a significantly fewer number of trajectories}, we applied \textit{MPPI} and \textit{log-MPPI} on a simulated cartpole system. 
\subsubsection{Simulation Setup:}\label{Simulation Setup:Cart-pole}
The main objective is to swing up and stabilize the cartpole for \SI{12}{\second}.
The cartpole dynamics model are taken from \cite{williams2017model} with assigning the same values to the system variables, 
while the pole length $l$ is set to \SI{1}{\metre} and the instantaneous running cost function is formulated as:
\begin{equation}
q(\mathbf{x})=\num{10}x^{2}+ \numprint{e3}(1+\cos (\theta))^{2}+2\dot{\theta}^{2} + 2\dot{x}^{2},
\end{equation}
where $x$ and $\dot{x}$ are the horizontal position and velocity of the cart, while $\theta$ and $\dot{\theta}$ denote the angle and angular velocity of the pole.   
For both control schemes, the simulations were performed with
a time prediction $t_p$ of \SI{2}{\second}, a control frequency of \SI{50}{\hertz} (sequentially, $N=100$), a \num{1000} sampled rollouts at each time-step $\Delta t$, an exploration variance $\nu$ of \num{1000}, an inverse temperature $\lambda$ of $0.07$, and a control weighting matrix $R$ of $\frac{\lambda}{2} \Sigma_{n}^{-\frac{1}{2}}$ with $\Sigma_{n} =0.0225$ for \textit{log-MPPI}.
The Savitzky-Galoy (\textit{SG}) convolutional filter, which is utilized for smoothing the control sequence $\left\{\mathbf{u}_{k}\right\}_{k=0}^{N-1}$ computed by Eq. \eqref{eq:mppi_optimal-control}, is applied with a quintic polynomial function, i.e., $n_{sg}=5$, and a window length $l_{sg}$ of $51$.

\begin{figure*}[!ht]
    \centering
    \subfloat[\centering \textit{MPPI} vs \textit{log-MPPI} for the same noise variance $\Sigma_\mathbf{u} = 0.283$, while $M=1000$]{\vspace*{-1pt}{\includegraphics[scale=1]{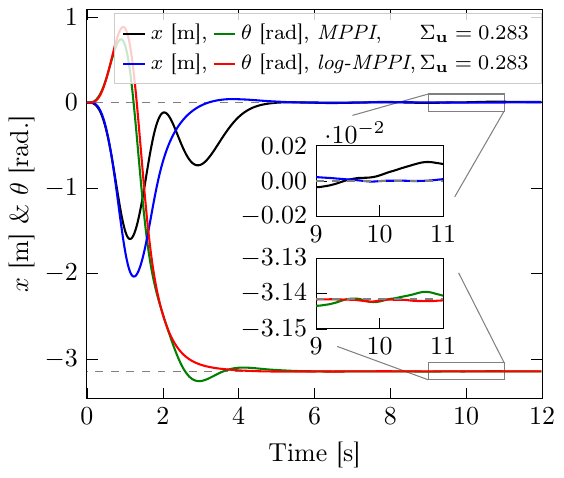}}\label{fig:cartpole_sameVar_283_K1000}}%
    \subfloat[\centering \textit{MPPI} vs \textit{log-MPPI} for the same noise variance $\Sigma_\mathbf{u} = 0.283$, while $M=5$ ]{\vspace*{-1pt}{\includegraphics[scale=1]{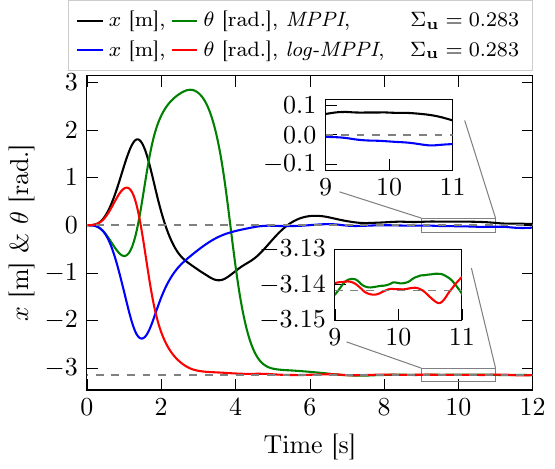}}\label{fig:cartpole_sameVar_283_K5}}%
    \subfloat[\centering Performance of \textit{MPPI} for higher values of $\Sigma_\mathbf{u}$, while $M=1000$ ]{\vspace*{-1pt}{\includegraphics[scale=1]{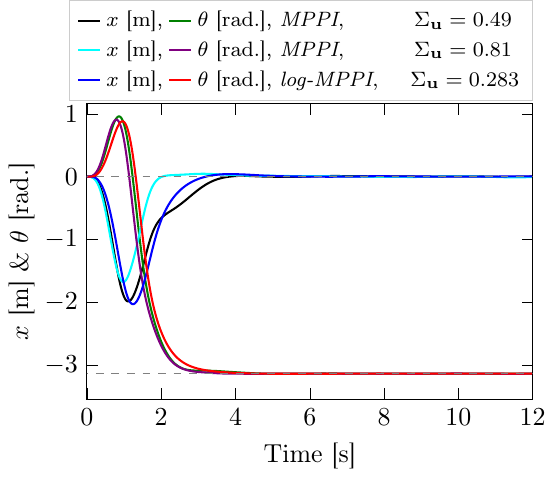}}\label{fig:cartpole_best_performance}}%
    \caption{Performance comparison of \textit{MPPI} and \textit{log-MPPI} for the cart–pole swingup task, considering: (a) the same sampling variance, namely, $\Sigma_\mathbf{u} = 0.283$, while $M=1000$, (b) the same sampling variance, i.e., $\Sigma_\mathbf{u} = 0.283$, while $M=5$, (c) higher variances in the case of \textit{MPPI}.}%
    \label{fig:cartpole-results}%
    \vspace*{-10pt}
\end{figure*}
\subsubsection{Simulation Results:}
We tested the robustness of our proposed algorithm by changing the noise variance $\Sigma_\mathbf{u}$, and number of sampled trajectories $M$, as illustrated in Fig.~\ref{fig:cartpole-results}. 
In Figs.~\ref{fig:cartpole_sameVar_283_K1000} and \ref{fig:cartpole_sameVar_283_K5}, the simulations are carried out considering two different values of $M$ (namely, $M=1000$ and $5$), while keeping the injected noise variance the same, namely, $\Sigma_\mathbf{u}=0.283$.
We can notice from Fig.~\ref{fig:cartpole_sameVar_283_K1000} that our control scheme achieves a slightly faster convergence; the cartpole converges to the desired configuration (i.e., $x=\SI{0}{\metre}$ and $\theta = \pi \,\si{\radian}$) within \SI{3.9}{\second} compared to \SI{4.8}{\second} when \textit{MPPI} is used.
Figure \ref{fig:cartpole_sameVar_283_K5} demonstrates that the impact of decreasing $M$ on the behavior of \textit{MPPI} is appreciably higher than its impact on \textit{log-MPPI}, as the former produced control input that ultimately leads to a higher transient overshoot of $\theta$ and higher positioning error (about \SI{7.6}{\centi\metre}).    
On the other side, \textit{log-MPPI} still performs well with a very slightly positioning error, without compromising its robustness level and convergence rate, which opens up a new avenue for sampling-based \textit{MPC} algorithm to be run on a standard \textit{CPU} instead of a \textit{GPU}{, with a fewer number of samples}.
\ihab{Furthermore, it is noteworthy that \textit{MPPI} can achieve similar or better performance of  \textit{log-MPPI} 
 by increasing $\Sigma_\mathbf{u}$ {at least 73\%}, as depicted in Fig.~\ref{fig:cartpole_best_performance}}\footnote{\ihab{Empirically, we observed that the lower the $R$, the better the performance. Accordingly, the behavior of \textit{MPPI} will be slightly worse if $R$ is assigned to a high value as that in \textit{log-MPPI}, which is $R=0.233$.}}. \ihab{However, assigning higher values might result in violating the system constraints if they are applied and added to the optimization control problem of the \textit{MPPI} algorithm.}

\subsection{Autonomous Navigation in Cluttered Environments}\label{Task1:Cluttered Environments}
With the aim of demonstrating the prospective advantages of our proposed \textit{log-MPPI} control strategy compared to the classical \textit{MPPI},
extensive simulations are conducted in 
goal-oriented \textit{AGV} autonomous navigation tasks in \textit{2D} cluttered environments.
\subsubsection{Simulation Setup:}\label{Simulation Setup:Cluttered Environments}
In this work, we consider the kinematics model of a
differential wheeled robot.
The kinematics equations that govern the motion of the robot is expressed as
\begin{equation}\label{eq:kinematics-model}
\left[\begin{array}{c}
\dot{x} \\
\dot{y} \\
\dot{\theta}
\end{array}\right]=\left[\begin{array}{cc}
\cos \theta & 0 \\
\sin \theta & 0 \\
0 & 1
\end{array}\right]\left[\begin{array}{c}
v \\
\omega
\end{array}\right] 
\Leftrightarrow \dot{\mathbf{x}} = \mathcal{R}(\theta)  \mathbf{u},
\vspace*{-2pt}
\end{equation}
where $\mathbf{x} = [{x}, {y}, \theta]^T \in \mathbb{R}^{3}$ represents the pose of the robot expressed in the world frame, 
$\theta$ is the rotation (or, heading) angle, 
the control $\mathbf{u} = [v,\omega]^T \in \mathbb{R}^{2}$ denote the linear and angular velocities of the robot. 

The parameters of both control strategies were set as follows: $t_p=\SI{5}{\second}$ (i.e., $N=250$), $M=2500$, $\nu =1200$, and $R=\lambda \Sigma_{n}^{-\frac{1}{2}}$. 
However, for \textit{MPPI}, the inverse temperature $\lambda$ and the control noise variance $\Sigma_\mathbf{u} = \operatorname{Diag}\left(\sigma_v^2, \sigma_w^2\right)$ (herein, $\Sigma_\mathbf{u} \equiv \Sigma_n$) are set to \num{0.572} and $\operatorname{Diag}\left(0.023, 0.028\right)$, respectively, while in the case of  \textit{log-MPPI} they are set to much lower values, namely, \num{0.169} and $\operatorname{Diag}\left(0.017, 0.019\right)$, respectively. 
In fact, those two hyperparameters were chosen based on the intensive simulations carried out in Tests \#1 and \#2 in Table~\ref{table:intensiveSimulation-Table}. 
It can be noticed that $\Sigma_\mathbf{u}$, in the case of \textit{log-MPPI}, is basically computed from a normal distribution with a variance of $\Sigma_{n} = \operatorname{Diag}\left(0.002, 0.0022\right)$.
For the \textit{SG} filter parameters, we set $n_{sg}$ and $l_{sg}$ to $3$ and $51$, respectively.
The \textit{real-time} execution of \textit{MPPI} and \textit{log-MPPI} is carried out on an NVIDIA GeForce GTX 1660 Ti laptop GPU, where all algorithms were written in Python and were implemented on a differential wheeled robot, namely, ClearPath Jackal robot, integrated with the Robot Operating System (\textit{ROS}) framework.

Within this work, trajectories are sampled on a \textit{GPU} using the discrete-time kinematics model given in Eq. \eqref{eq:kinematics-model}, where the state-dependent cost function of the \textit{2D} navigation task is simply formulated as
\begin{equation}\label{eq:state-dep-cost-function}
q(\mathbf{x})= q_{\text{state}}(\mathbf{x}) + q_{\text{obs}}(\mathbf{x}),
\end{equation}
where $q_{\text{state}}(\mathbf{x}) = (\mathbf{x}-\mathbf{x}_{f})^{\mathrm{T}} Q (\mathbf{x}-\mathbf{x}_{f})$ is a quadratic cost function utilized for enforcing the robot current state $\mathbf{x}$ to reach its desired state $\mathbf{x}_{f}$, and $Q = \operatorname{Diag}(5,5,2) \;\forall v_\text{des}\leq \SI{1}{\metre/\second}$, otherwise $Q = \operatorname{Diag}(2.5,2.5,2)$. $q_{\text{obs}}(\mathbf{x}) = 10^7C_{\text{crash}}$ heavily penalizes trajectories that collide with obstacles, where $C_{\text{crash}}$ is a Boolean variable that indicates the collision with obstacles.   
\subsubsection{Simulation Scenarios:}
We considered four various scenarios for evaluating the performance of the proposed control framework in cluttered environments. 
In \textit{Scenario \#1}, the intensive simulations (namely, $\mathcal{N}_{T} = 100$ tasks) are carried out by taking into account different values of $\lambda$ and $\Sigma_{\mathbf{u}}$, with the aim of (i) choosing the best sets of hyperparameters that respect the control constraints, then (ii) assessing the performance in the following scenarios.
We choose $\lambda \in [0.01, 10]$ and $\Sigma_{\mathbf{u}} \in [0.0004, 0.16]$, while a cluttered environment, with obstacles placed \SI{2}{\metre} away, has been used for assessing the performance.
In the last three scenarios, we randomly generated three different types of forests, each type has \num{50} forests, i.e., $\mathcal{N}_{T} = 50$ tasks, and each forest represents a $\SI{50}{\metre} \times \SI{50}{\metre}$ cluttered environment. 
In the first type (i.e., \textit{Scenario \#2}), the obstacles were, on average, \SI{1.5}{\metre} apart (namely, $d^{\text{obs}}_{\min}=\SI{1.5}{\metre}$), whilst in the second (\textit{Scenario \#3}) and third (\textit{Scenario \#4}), they placed \SI{2}{\metre} and \SI{3}{\metre} away, respectively. 
For the first three scenarios, the maximum desired velocity $v_\text{des}$ of the robot is set to \SI{1.5}{\metre/\second}, while in the latter it is allowed for the robot to navigate with its maximum velocity which is \SI{2}{\metre/\second}.

\subsubsection{Performance Metrics:}
To achieve a fair performance comparison of the two control schemes: (i) first, in all simulations, the robot has to reach the same desired pose, namely, $\mathbf{x}_{f}= [50,50,0]^T$, from the predefined initial pose $\mathbf{x}_{0}= [0,0,0]^T$ (in [\si{\metre}], [\si{\metre}], [\si{\deg}]); (ii) second, we define a set of metrics so as to assess the overall performance such as the number of successful tasks $\mathcal{S}_{T}$, success rate $\mathcal{S}_{R}$, average \ihab{robot} trajectory length $l_{\text{av}}$ to reach $\mathbf{x}_{f}$ from $\mathbf{x}_{0}$, number of successful tasks \ihab{
with a shorter route (i.e., robot trajectory) towards the goal} $\mathcal{N}_{l_{\text{min}}}$, and average traveling speed $v_{\text{av}}$.
The task is considered to be successful if the robot successfully reaches the desired goal without colliding with obstacles.
\ihab{Note that $l_{\text{av}}$, $\mathcal{N}_{l_{\text{min}}}$, and $v_{\text{av}}$ are only computed for 
successful tasks $\mathcal{S}_{T}$ that are successfully completed by both control schemes.} 

\begin{table}[!ht]
\small\addtolength{\tabcolsep}{-3.50pt} 
\setlength\extrarowheight{1pt}
\caption{Overall performance of the two control schemes, where the gray cells represent better results.} 
\centering
\begin{tabular}{|c|| c | c| c| c| c| c| c|}
\hline
 Test  & Scheme  &  $\mathcal{N}_{T}$  & $\mathcal{S}_{T}$ & $\mathcal{S}_{R}$ [\%] & $l_{\text{av}}$ [\si{\metre}] ($\mathcal{N}_{l_{\text{min}}}$)& $v_{\text{av}}$ [\si{\metre/\second}]\\
 \hline  
 \hline
 \multicolumn{7}{|c|}{\textit{\textbf{Scenario \#1:}} $v_{\text{des}} = \SI{1.5}{\metre/\second}$ \&  $d^{\text{obs}}_{\min}=\SI{2}{\metre}$} \\
 \hline
\#1 & \textit{MPPI} & 100 & 48 & 48 & $-$ & $-$ \\
\#2 & \textit{log-MPPI} & 100 & \cellcolor{gray!20} 61 &  \cellcolor{gray!20} 61 & $-$ & $-$ \\
\#$\overline{1}$ & \textit{MPPI} & 60 & 21 & 35 & $-$ & $-$ \\
\#$\overline{2}$ & \textit{log-MPPI} & 60 &  \cellcolor{gray!20} 43 &  \cellcolor{gray!20} 71.7 & $-$ & $-$ \\
 \hline \hline
 \multicolumn{7}{|c|}{\textit{\textbf{Scenario \#2:}} $v_\text{des} = \SI{1.5}{\metre/\second}$ \& $d^{\text{obs}}_{\min}=\SI{1.5}{\metre}$} \\
 \hline
 \#3 & \textit{MPPI} & 50 & 42 & 84 & 76.52 (9/40) & $1.27 \pm 0.18$  \\
\#4 & \textit{log-MPPI} & 50 & \cellcolor{gray!20} 48 & \cellcolor{gray!20} 96 & \cellcolor{gray!20} 75.19 (31/40) & \cellcolor{gray!20} $1.34 \pm 0.12$ \\ 
 \hline
 \hline
 \multicolumn{7}{|c|}{\textit{\textbf{Scenario \#3:}} $v_\text{des} = \SI{1.5}{\metre/\second}$ \& $d^{\text{obs}}_{\min}=\SI{2}{\metre}$} \\
 \hline 
 \#5 & \textit{MPPI} & 50 & 46 & 92\% & 76.19 (13/46) & $1.32 \pm 0.14$ \\
\#6 & \textit{log-MPPI} & 50 & \cellcolor{gray!20} 50 & \cellcolor{gray!20} 100 & \cellcolor{gray!20} 75.29 (33/46) & \cellcolor{gray!20} $1.33 \pm 0.11$ \\ 
 \hline
 \hline
 \multicolumn{7}{|c|}{\textit{\textbf{Scenario \#4:}} $v_\text{des} = \SI{2}{\metre/\second}$ \& $d^{\text{obs}}_{\min}=\SI{3}{\metre}$} \\
 \hline
 \#7 & \textit{MPPI} & 50 & \cellcolor{gray!20} 50 & \cellcolor{gray!20} 100 & 72.17 (21/50) & $1.82  \pm 0.039$ \\ 
\#8 & \textit{log-MPPI} & 50 & \cellcolor{gray!20} 50 & \cellcolor{gray!20} 100 & \cellcolor{gray!20} 72.09 (29/50) & \cellcolor{gray!20} $1.84  \pm 0.037$
\\ 
 \hline
 \end{tabular}
\label{table:intensiveSimulation-Table}
 \vspace*{-10pt} 
\end{table}
\subsubsection{Simulation Results:}
The general performance of our proposed control schemes are summarized in Table~\ref{table:intensiveSimulation-Table}, considering the four scenarios defined previously and controllers' parameters given in Section~\ref{Simulation Setup:Cluttered Environments}. 
It is worthy to notice in \textit{Scenario \#1} (i.e., Tests \#1 and \#2), where different values of $\lambda$ and $\Sigma_{\mathbf{u}}$ are considered, that \textit{log-MPPI} significantly outperforms \textit{MPPI} as its success rate $\mathcal{S}_{R}$ is 
noticeably 
higher than that in \textit{MPPI}, 
especially for the first $60$ tasks (i.e., $\mathcal{N}_{T}=60$) where lower values are assigned to $\Sigma_{\mathbf{u}}$ (see Tests \#$\overline{1}$ and \#$\overline{2}$)\footnote{The motive behind considering the first 60 tasks (namely, Tests \#$\overline{1}$ and \#$\overline{2}$) is that we empirically observed that assigning higher values to $\Sigma_{\mathbf{u}}$ increases the possibility of violating the control constraints.}.  
\ihab{In practice, this clearly indicates that \textit{log-MPPI} is largely compatible with a wide range of acceptable parameters values, reducing the time taken for fine-tuning those parameters that play an important role in determining the behavior of sampling-based \textit{MPC} scheme.}
For \textit{Scenario \#2} (i.e., Tests \#3 and \#4), where $d^{\text{obs}}_{\min}=\SI{1.5}{\metre}$, it can be clearly noticed that our method experimentally exhibits better performance not only due to its higher success rate ($\mathcal{S}_{R}=96\%$), but also due to the fact that: (i) $l_{\text{av}}$ is lightly shorter (roughly, \SI{1.33}{\metre} shorter than that for \textit{MPPI}), (ii) $\mathcal{N}_{l_{\text{min}}}$ is quite higher (totally, \num{31} tasks compared to \num{9} for \textit{MPPI}), and (iii) $v_{\text{av}}$ is slightly better and closer to $v_{\text{des}}$ with a very low standard deviation.
Similarly, in the remaining tests with high values of $d^{\text{obs}}_{\min}$, \textit{log-MPPI} performs well with a high capability of successfully complete all given tasks while avoiding obstacles; thanks to the \textit{NLN} distribution policy that provides more flexible and efficient trajectories, \ihab{we 
ensure} a much better exploration of the state-space of the given system \ihab{with more than 30\% reduction in the injected noise variance} and \ihab{
reduce} the risk of getting stuck in local minima \ihab{when \textit{MPPI} is employed}.

\begin{figure}%
  \centering
  \begin{minipage}[t]{.55\linewidth}
    \subcaptionbox{\centering Robot trajectories generated by \textit{MPPI} and  \newline \textit{log-MPPI}\label{fig:obs_map}; red dots represent random obstacles 
    }
      {\hspace*{-5pt}\includegraphics[scale=1]{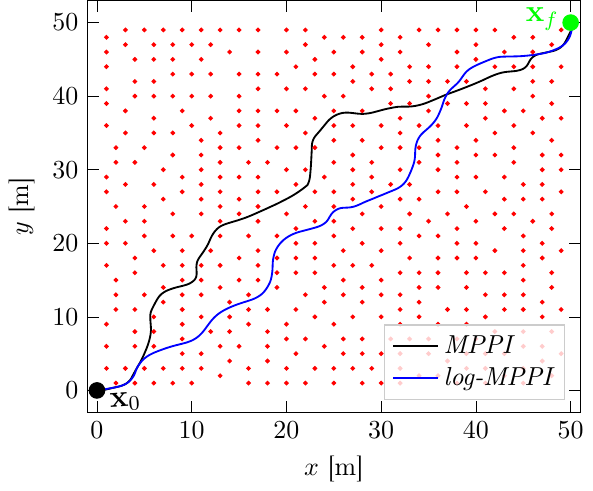}}%
  \end{minipage}%
  \hfill
  \begin{minipage}[b]{.34\linewidth}
    \subcaptionbox{Linear velocity (\textit{MPPI)}\label{fig:vel-mppi}}
       {\includegraphics[scale=1]{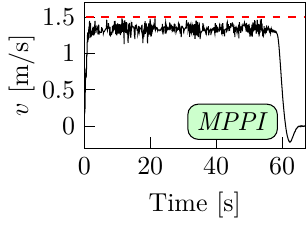}}
    \subcaptionbox{\centering Linear velocity (\textit{log-MPPI)}\label{fig:vel-logmppi}}
      {\includegraphics[scale=1]{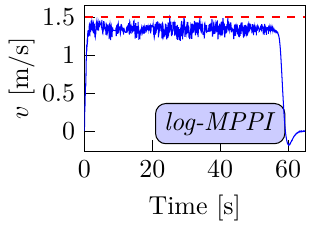}}%
  \end{minipage}%
  \caption
    {%
      Behavior of \textit{MPPI} and \textit{log-MPPI} in a $\SI{50}{\metre} \times \SI{50}{\metre}$ cluttered environment, where obstacles were \SI{2}{\metre} apart.%
      \label{fig:example1}%
    }%
    \vspace*{-11pt}
\end{figure}

In Fig. \ref{fig:example1}, we show an example of the robot trajectories generated by \textit{MPPI} and \textit{log-MPPI} in a cluttered environment
, where $d^{\text{obs}}_{\min}=\SI{2}{\metre}$, i.e., \textit{Scenario \#3}.
We can clearly observe that although both control schemes achieve successfully collision-free navigation through the cluttered environment with an average traveling speed $v_{\text{av}}$ of \SI{1.33}{\metre/\second} \ihab{which respects the control constraints as the robot linear velocity $v\leq v_{\text{des}}=\SI{1.5}{\metre/\second}$} as shown in Figs.~\ref{fig:vel-mppi} and \ref{fig:vel-logmppi}, \textit{log-MPPI} provides a shorter route towards the goal as shown in Fig.~\ref{fig:obs_map}. More precisely, the length of the robot trajectory $l$ in the case of \textit{log-MPPI} is \SI{77.8}{\metre} compared to \SI{79.92}{\metre} for the classical \textit{MPPI}.
\vspace*{-13pt}
\subsection{Autonomous Navigation in Unknown Environments}\label{Task2: Unknown Environments}
\begin{figure}[!th]%
    \vspace*{-5pt}
    \centering
    \subfloat[Gazebo forest-like environment]{\vspace*{1pt}{\includegraphics[height=1.in, width=.5\columnwidth]{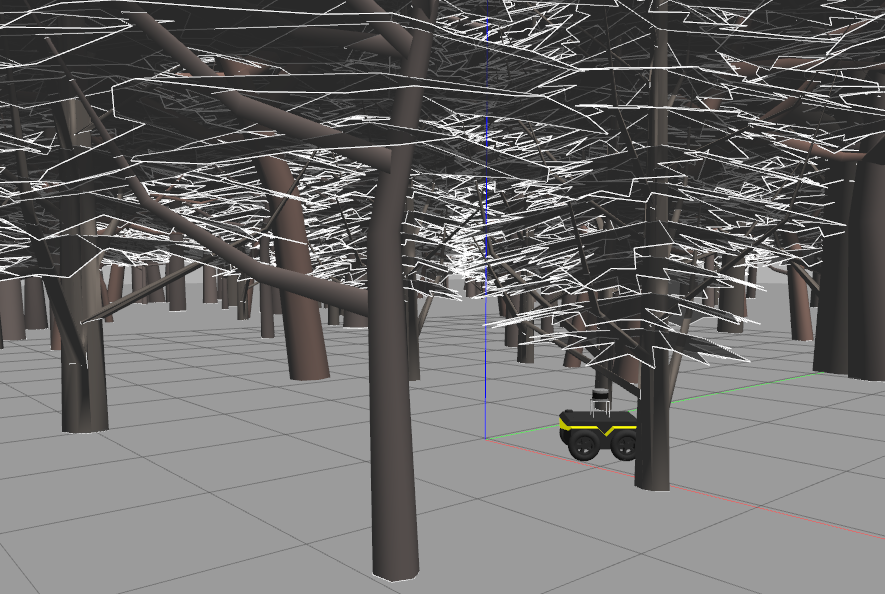}}\label{fig:forst-gazebo}}%
    \;\!\!
    \subfloat[Rviz costmap visualization]{\vspace*{1pt}{\includegraphics[height=1.in, width=.46\columnwidth]{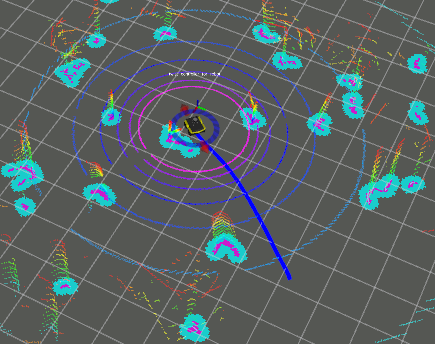}}\label{fig:forst-costmap}}%
    \caption{Snapshot of (a) our Jackal robot 
    located in a forest-like environment, and (b) the corresponding \textit{2D} local costmap.
    \vspace*{-3pt}
    }
    \label{fig:forst-gazebo-rviz}%
\end{figure}

In Section \ref{Task1:Cluttered Environments}, where autonomous navigation in cluttered environments is performed, it is assumed that the costmap that represents the environment is priorly known, limiting the applicability of the control schemes as unknown or partially observed environments are the most dominant in robotics applications \cite{mohamed2020model}.
To this end, a  \textit{2D} costmap created by the \textit{costmap\_2d ROS} package is utilized for storing information about the robot's surrounding obstacles \cite{costmap2016} (see Fig.~\ref{fig:forst-costmap}). 
It employs the sensor data acquired from the environment to build a  \textit{2D} or \textit{3D} occupancy grid of the data, where each cell of the occupancy is typically expressed as \textit{occupied, free, or unknown}; in our case, a \textit{2D} occupancy grid map is sufficient for \textit{2D} robot navigation. 
Thereafter, this occupancy grid is utilized as a local costmap to be fed directly into the sampling-based \textit{MPC} algorithm, for achieving collision-free navigation in either static or dynamic unknown environments.
\subsubsection{Simulation Setup:}\label{Simulation Setup:Unknown Environments} 
We considered the same simulation setup previously described in Section \ref{Simulation Setup:Cluttered Environments}, where the collision indicator function $q_{\text{obs}}(\mathbf{x})$, given in (\ref{eq:state-dep-cost-function}), is herein computed based on the local costmap (i.e., \textit{2D} grid map) built by the robot on-board sensor; in this work, the Clearpath Jackal robot is endowed with a Velodyne VLP-16 LiDAR sensor. 
The size of the robot-centered \textit{2D} grid map is set to $\SI{240}{\cell} \times \SI{240}{\cell}$ with a resolution (grid size) of \SI{0.05}{\metre/\cell}.
\subsubsection{Simulation Scenarios:} For the benchmark, two types of $\SI{50}{\metre} \times \SI{50}{\metre}$ forest-like maps in Gazebo environment are utilized, as depicted in Fig.~\ref{fig:forst-gazebo}. 
The first type (namely, \textit{Forest \#1}) contains tree shaped obstacles with a density of \SI{0.1}{\trees/\square \metre}, while the latter (i.e., \textit{Forest \#2}) with a density of \SI{0.2}{\trees/\square \metre}. 
In the case of \textit{Forest \#1}, $v_\text{des}$ is set to \SI{2}{\metre/\second}, while it is reduced to \SI{1.5}{\metre/\second} in the case of \textit{Forest \#2}.
Another scenario (namely, \textit{Corridor \#1}) is considered in which the robot navigates along a $\SI{6}{\metre} \times \SI{20}{\metre}$ corridor, with $v_\text{des}=\SI{1.5}{\metre/\second}$, in the presence of \num{8} pedestrians, each pedestrian holding a maximum velocity of $v_\text{ref}=\SI{0.3}{\metre/\second}$.

\subsubsection{Performance Metrics:}
Here, we conduct a comparison between the two control schemes in the aspect of the number of collisions $N_{\text{crash}}$, average trajectory length $l_{\text{av}}$, average execution time per iteration $t_{\text{mppi}}$ of the control algorithm. The desired poses (in ([\si{\metre}], [\si{\metre}], [\si{\deg}])) are defined as follows: $G_1= [0,0,0]^T, G_2= [20,20,45]^T, G_3= [-18,2,0]^T, G_4= [20,-21, 90]^T$, then $G_5= [20,20,0]^T$. For the sake of simplicity, for \textit{Forest \#2}, the robot navigates autonomously from $G_1$ to only $G_3$, then stops.

\subsubsection{Simulation Results:}
Table~\ref{table:Comparison-unknowenviroments} summarizes the performance statistics of the two proposed control strategies in \textit{Forest \#1} and \textit{Forest \#2}, considering 10 trials for each. 
As anticipated, the obtained results demonstrate that \textit{log-MPPI} has a more flexible and efficient trajectories sampling distribution policy, resulting in (i) reducing the probability of getting stuck in a local minima (e.g., in \textit{Forest \#2}, $N_{\text{crash}} = 1$ compared to $N_{\text{crash}}=7$ when \textit{MPPI} is used), and (ii) improving the quality of the generated trajectory as $l_{\text{av}}$ is appreciably shorter, especially in \textit{Forest \#2}. 
Furthermore, we can emphasize that both control schemes guarantee a \textit{real-time} performance (as $t_{\text{mppi}} < \SI{20}{\milli\second}$), showing the superiority of the sampling-based \textit{MPC} algorithm to be deployed with \textit{2D} grid maps without adding any additional complexity to the optimization problem.

For a \textit{2D} grid-based navigation in the dynamic environment (namely, \textit{Corridor \#1}), the simulation results demonstrate that the autonomous vehicle successfully avoids moving agents, as shown in the supplementary video. However, we empirically noticed that the more the deployed agents, the noisier the \textit{2D} costmap, increasing the risk of being trapped in local minima.
\begin{table}[!ht]
\vspace*{-12.5pt}
\caption{Performance comparisons of \textit{MPPI} and \textit{log-MPPI}.}
\vspace{-5pt}
\begin{center}
\small\addtolength{\tabcolsep}{-5.5pt} 
 \begin{tabular}{|l||c|c|c|c|}
 \hline
& \multicolumn{2}{c|}{\textit{\textbf{Forest \#1}}} & \multicolumn{2}{c|}{\textit{\textbf{Forest \#2}}}\\ 
 \cline{2-5}
 \multirow{-2}{*}{ \hspace*{-2pt}\textit{Indicator}}  & \multicolumn{1}{c|}{\textit{MPPI}} & \multicolumn{1}{c|}{\textit{log-MPPI}} & \multicolumn{1}{c|}{\textit{\textit{MPPI}}} & \multicolumn{1}{c|}{\textit{\textit{log-MPPI}}}\\
 \hline\hline
 $N_{\text{crash}}$    & 2 & \cellcolor{gray!20} 0 & 7 &   \cellcolor{gray!20} 1 \\
 $l_{\text{av}}$ {[\si{\metre}]}& $158.71\pm 1.54$& \cellcolor{gray!20}$157.91\pm 0.54$& $76.12\pm 3.31$& \cellcolor{gray!20} $72.64\pm0.80$\\
 $t_{\text{mppi}}$ {[\si{\milli\second}]} & \cellcolor{gray!20} $ 13.72 \pm 1.34$ & $13.98  \pm 0.68$          & $13.70  \pm 4.26$  & \cellcolor{gray!20} $13.47 \pm 1.46$ \\
 \hline
\end{tabular}
\end{center}
\label{table:Comparison-unknowenviroments}
\end{table}
\vspace{-12pt}
\section{Real-World Demonstration}\label{Real-World Demonstration}
 We experimentally demonstrate the applicability of \ihab{
 the} \ihab{
 proposed} control \ihab{
 strategies} for achieving a \textit{2D} grid-based collision-free navigation in an unknown indoor cluttered environment. 
\subsubsection{Experimental Setup:}\label{Experimental Setup:real-world Environment} 
The simulation setup formerly described in Sections \ref{Simulation Setup:Cluttered Environments} and \ref{Simulation Setup:Unknown Environments} is also employed for the experimental validation. 
However, as the indoor environment size is tiny compared to that used for the simulation scenarios, we set $v_\text{des} = \SI{0.75}{\metre/\second}$ and $t_p =\SI{8}{\second}$, while the \textit{2D} grid map size is decreased to half of its nominal value (i.e., $\SI{120}{\cell} \times \SI{120}{\cell}$) to ensure a \textit{real-time} implementation of the control strategies.
Our experimental platform is a fully autonomous Clearpath Jackal robot equipped with a 16-beam Velodyne LiDAR sensor utilized for (i) generating the local costmap, and (ii) estimating the robot's pose using \textit{LOAM} \cite{zhang2014loam}.
\subsubsection{Validation Environment:}
A $\SI{7}{\metre} \times \SI{5.5}{\metre}$ indoor cluttered environment with random boxes as obstacles is utilized for experimental validation, as depicted in Fig.~\ref{fig:real-world-enviro}, where obstacles were, on average, \SI{1.3}{\metre} apart. Herein, the desired poses (in ([\si{\metre}], [\si{\metre}], [\si{\deg}])) are given as follows: $G_1= [0,0,0]^T, \ihab{
G_2= [7,5.1, 45]^T}, G_3= [5.5,1.5, 0]^T, G_4= [0,3, 45]^T$, $G_2$, then $G_1$ (see Fig.~\ref{fig:real-world-enviro}).

\subsubsection{Experimental Results:}\label{Experimental results:real-world Environment} 
\begin{figure}%
{
    \centering
    \subfloat{{\includegraphics[height=0.85in, width=.24\columnwidth]{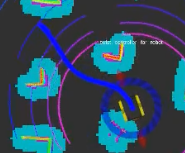}}}%
    \;\!\!
    \subfloat{{\includegraphics[height=0.85in, width=.24\columnwidth]{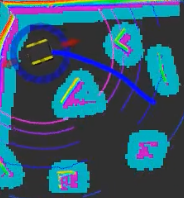}}}%
     \;\!\!
    \subfloat{{\includegraphics[height=0.85in, width=.24\columnwidth]{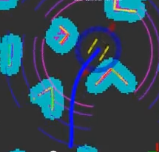}}}%
     \;\!\!
    \subfloat{{\includegraphics[height=0.85in, width=.24\columnwidth]{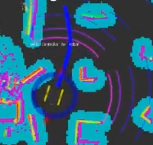}}}%
    \vspace*{3pt}
    \caption{{Planned trajectory by \textit{log-MPPI} using the 2D costmap \ihab{previously described in Fig.~\ref{fig:indoor-costmap}} (left to right) at $t=\SI{8}{\second}\, (\!G_1\rightarrow G_2\!)$, $t=\SI{20}{\second}\, (\!G_2\rightarrow G_3\!)$, $t=\SI{31}{\second}\, (\!G_3\rightarrow G_4\!)$, and 
   $t=\SI{47}{\second}\, (\!G_4\rightarrow G_2\!)$.}}
    \label{fig:_Experimenatl-snap}
    }
    \vspace*{-12pt}
\end{figure}

The performance statistics for three trials in our indoor environment is summarized in Table~\ref{table:real-world-results}.
We can clearly observe, for all trials, that \ihab{
both} control scheme\ihab{s} provide \textit{real-time} collision-free navigation, since $N_{\text{crash}}=0$ and $t_{\text{mppi}} < \SI{20}{\milli\second}$, in the cluttered environment with an average traveling speed $v_{\text{av}}$ of \ihab{
\SI{0.56}{\metre/\second}}, regardless of the limited perception range. \ihab{In addition,
the quality of the generated trajectories by \textit{log-MPPI} is considerably better than that generated by \textit{MPPI}, as $l_{\text{av}}$ is noticeably shorter especially considering the scale of the environment and the density of random obstacles in it.} {Figure~\ref{fig:_Experimenatl-snap} shows a snapshot of the collision-free and \ihab{predicted optimal} trajectories generated by \textit{log-MPPI} \ihab{at different time instants while the robot navigates to the desired poses}}. More details about the experimental results are included in this video: \url{https://youtu.be/bLrQWYLgocw}.
\begin{table}[!ht]
 \vspace*{-3pt}
\caption{
\ihab{
Performance statistics of the two control strategies.
}}
\vspace{-6pt}
\begin{center}
\small\addtolength{\tabcolsep}{-2pt} 
 \begin{tabular}{|c||c|c|c|c|c|}
 \hline
Scheme & $N_{\text{crash}}$  & $l_{\text{av}}$ [\si{\metre}] & $v_{\text{av}}$ [\si{\metre/\second}] & $t_{\text{mppi}}$ {[\si{\milli\second}]}\\
\hline
\textit{MPPI} & \cellcolor{gray!20} 0 & $40.25  \pm 0.13$    & $0.55  \pm 0.13$  & $11.43 \pm 0.24$ \\
\textit{log-MPPI} & \cellcolor{gray!20} 0 & \cellcolor{gray!20} $38.95  \pm 0.17$    & \cellcolor{gray!20} $0.57  \pm 0.14$  & \cellcolor{gray!20} $11.18 \pm 0.08$ \\
 \hline
\end{tabular}
\end{center}
\label{table:real-world-results}
\end{table}
\vspace*{-12pt}
\section{Conclusion \ihab{and Future Work}}\label{sec:conclusion}
In this work, we proposed an extension to the classical \textit{MPPI} algorithm (namely, \textit{log-MPPI}) in which the control input updates  $\delta \mathbf{u}_{k}$ are sampled from the normal log-normal (\textit{NLN}) mixture distribution, rather than from Gaussian distribution. 
We also presented a sampling-based \textit{MPC} framework for collision-free navigation in either static or dynamic unknown cluttered environments, by directly integrating the occupancy grid as a local costmap into the sampling-based \textit{MPC} algorithm. 
We empirically demonstrated \ihab{
} that the trajectory samples generated by \textit{log-MPPI} are more flexible and efficient than the ones generated by \textit{MPPI}, with a more than 30\% reduction in the injected noise variance $\Sigma_{\mathbf{u}}$ when \textit{MPPI} is employed.
This subsequently results in exploring much better the state-space of the controlled system and reducing the risk of getting stuck in local minima. 
We demonstrated in real-world environment the possibility of feeding directly the local costmap into the optimal control problem without adding any additional complexity to the control problem, as well as ensuring a \textit{real-time} performance of the proposed control strategy. 
\ihab{In the future, we plan to implement our control scheme on standard \textit{CPUs} rather than \textit{GPUs}, aiming to reduce the computational burden.
Furthermore, we will explore methods for vanishing the possibility of getting stuck in local minima and studying the \textit{theoretical} stability of sampling-based \textit{MPC}.}


\section*{Acknowledgement}
The authors would like to thank Grady Williams, Ziyi Wang, and Evangelos Theodorou for the fruitful discussions 
for improving the work.

\bibliographystyle{IEEEtran}
\bibliography{references}            

\appendices

\section{
Analysis of the New Sampling Strategy on \textit{MPPI}}\label{Appendix-dynamics-lognormal-normal}
In this appendix, we provide a brief interpretation for the trajectory rollout behavior using the proposed sampling method.
Let us examine the effect on the dynamics due to the change of noise.
The original dynamics of \textit{MPPI} reads (i.e., Eq. (53) in \cite{williams2018robust}, ignored constant) 
\begin{align}
    \!\!\!\!d\bm{x}_t\!=\!\bm{f}(\bm{x}_t, t)\,dt \!+ \!\bm{G}(\bm{x}_t, t)\bm{u}(\bm{x}_t, t)\,dt \!+ \!\bm{G}(\bm{x}_t, t)\,d\bm{B}(t),
\end{align}
where $\bm{B}(t)$ is Brownian motion.

In the discrete version, if $\bm{\epsilon}$ in the sampling is replaced by $\bm{Z}=\bm{\epsilon Y}=\bm{\epsilon} e^{\bm{W}}$, where $\bm{W}$ is an independent Gaussian vector of $\bm{\epsilon}$, we may consider this randomness as a new disturbance to the original dynamics. Although it is difficult to obtain an exact dynamics corresponding to the proposed method, we may use the following approximation. 
Assume that the original $d\bm{B}(t)$ is replaced by $d\left(\bm{B}(t) e^{\mu_{ln}t+\sigma_{ln}\bm{B}_1(t)}\right)$, where $\bm{B}_1(t)$ is a standard Brownian motion, independent of $\bm{B}(t)$. To simplify notation,
denote $\mu_{ln}t+\sigma_{ln}\bm{B}_1(t)$ by $\bm{W}_1$, and $\kappa\triangleq \frac{1}{2}\sigma_{ln}^2 + \mu_{ln}$.
By Ito's formula, we can get the following computation for $ d\left(\bm{B}(t) e^{\bm{W}_1(t)}\right)$: 
\begin{align}
   &e^{\bm{W}_1(t)}d\bm{B}(t) + \bm{B}(t)de^{\bm{W}_1(t)} + d[\bm{B}(t), e^{\bm{W}_1(t)}]\nonumber\\
    &=e^{\bm{W}_1\!(t)}d\bm{B}(t) \!+\! \!\bm{B}(t)\!\left(\!\sigma_{ln}e^{\bm{W}_1\!(t)}d\bm{B}_1\!(t)  
    \!+\!\kappa e^{\bm{W}_1\!(t)}\!dt \!\right)\!.
\end{align}
Thus, the sampling dynamics can be viewed as a modified one
\begin{align}
    &d\bm{x}_t=\bm{f}(\bm{x}_t, t)\,d t + \bm{G}(\bm{x}_t, t)\bm{u}(\bm{x}_t, t)\,dt \nonumber\\
    & + \bm{G}(\bm{x}_t, t)\, d\left(\bm{B}(t) e^{\bm{W}_1(t)}\right)\nonumber\\
    &=\bm{f}(\bm{x}_t, t)\,d t + \bm{G}(\bm{x}_t, t)\left(\bm{u}(\bm{x}_t, t) + 
    \kappa\bm{B}(t)
    e^{\bm{W}_1(t)}\right)dt\nonumber\\ 
    &+\bm{G}(\bm{x}_t, t)\left( e^{\bm{W}_1(t)}d\bm{B}(t) + \sigma_{ln}\bm{B}(t)e^{\bm{W}_1(t)}d\bm{B}_1(t)\right).
\end{align}
We have two observations. 
First, the drift term (the term with $dt$) is modified to 
$\bm{f}(\bm{x}, t) + \bm{G}(\bm{x}, t)\bm{u}(\bm{x}_t, t)+\kappa\bm{G}(\bm{x}, t)\bm{B}(t) e^{\bm{W}_1(t)}$.
This can be thought of a random drift term, compared to the deterministic counterpart in the original dynamics. But the mean of the drift term remains the same with the original one. Since the drift term can be viewed as the ``trend" of the path for each sample path, the proposed scheme has more diverse ``trends" of the trajectories than the original one. 
The new sampled trajectories turn to spread out much more than that in the original dynamics, indicating that this scheme can explore more spaces than the original one. 
This leads to a more efficient sampling scheme than the normal distribution, even with the similar variance.
Second, the noise term can be much more flexible to tune the variance of the sampling trajectories as it contains more parameters. 


\end{document}